\documentclass[final]{cvpr}

\usepackage{times}
\usepackage{epsfig}
\usepackage{graphicx}
\usepackage{amsmath}
\usepackage{amssymb}

\usepackage[ruled,vlined]{algorithm2e}
\SetAlFnt{\small}
\usepackage{siunitx}
\usepackage{booktabs}
\usepackage{makecell}
\usepackage{enumitem} % reduce space in bullet-point list
\usepackage{adjustbox}
\usepackage{subfig}

\usepackage{array}
\usepackage{balance}
\newcolumntype{P}[1]{>{\centering\arraybackslash}p{#1}}
\newcolumntype{M}[1]{>{\centering\arraybackslash}m{#1}}
\usepackage{color}

\newcommand{\Fig}{Fig.~}
\newcommand{\Tab}{Table~}

\def\cvprPaperID{36} % *** Enter the CVPR Paper ID here
\def\confYear{CVPRW 2021}
\newcommand{\rone}[1]{{\color{black}#1}}
\newcommand{\rtwo}[1]{{\color{black}#1}}

\newcommand\MYhyperrefoptions{bookmarks=false,bookmarksnumbered=true,pagebackref=true,breaklinks=true,
pdfpagemode={UseOutlines},plainpages=false,pdfpagelabels=true,
colorlinks=true,citecolor={black},
pdftitle={How to Calibrate Your Event Camera},%
pdfsubject={Computer Vision, Event Vision, Robotics, Neuromorphic Engineering},%
pdfauthor={M. Muglikar, M. Gehrig, D. Gehrig, D. Scaramuzza},%
pdfkeywords={Event Cameras, Low Latency, High Dynamic Range, Low Power}}%
\usepackage[\MYhyperrefoptions,pdftex]{hyperref}

\definecolor{somegray}{rgb}{0.5, 0.5, 0.5}
\newcommand{\darkgrayed}[1]{\textcolor{somegray}{#1}}
\makeatletter
\newcommand*\titleheader[1]{\gdef\@titleheader{#1}}
\AtBeginDocument{%
  \let\st@red@title\@title
  \def\@title{%
    \vskip-3em
    \bgroup\normalfont\large\centering\@titleheader\par\egroup
    \vskip1.5em\st@red@title}
}
\makeatother

\titleheader{\darkgrayed{This paper has been accepted for publication at the\\
IEEE Conference on Computer Vision and Pattern Recognition Workshops (CVPRW), Nashville, 2021.
\copyright IEEE}}

\title{How to Calibrate Your Event Camera}

\begin{document}
\author{Manasi Muglikar\thanks{Equal contribution}\qquad Mathias Gehrig$^{*}$ \qquad Daniel Gehrig \qquad Davide Scaramuzza\\
Dept. Informatics, Univ. of Zurich and \\
Dept. of Neuroinformatics, Univ. of Zurich and ETH Zurich\\
}

\maketitle



%\and

\begin{abstract}
We propose a generic event camera calibration framework using image reconstruction. Instead of relying on blinking LED patterns or external screens, we show that neural-network--based image reconstruction is well suited for the task of intrinsic and extrinsic calibration of event cameras. The advantage of our proposed approach is that we can use standard calibration patterns that do not rely on active illumination. Furthermore, our approach enables the possibility to perform extrinsic calibration between frame-based and event-based sensors without additional complexity. Both simulation and real-world experiments indicate that calibration through image reconstruction is accurate under common distortion models and a wide variety of distortion parameters.
\end{abstract}

\section*{Multimedia Material} 
The project's code is available at \\ \url{https://github.com/uzh-rpg/e2calib}
\section{Introduction}
\begin{figure}[t]
    \centering
    \includegraphics[trim={0cm 1.0cm 1cm 0cm},clip,width=\linewidth]{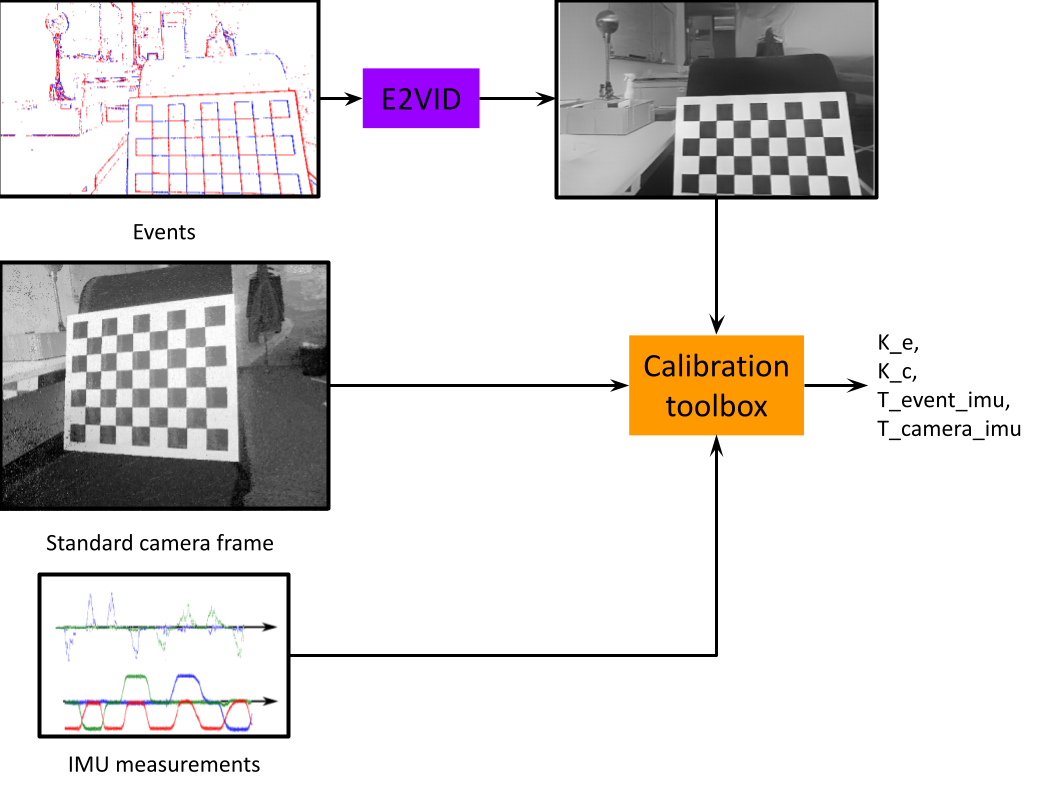}
    \vspace{-1ex}
    \caption{Our approach for event camera calibration of a hybrid multi-sensor setup consisting of an event camera, a standard camera and an IMU.
    The asynchronous and sparse event stream is passed to the network E2VID\cite{Rebecq19pami} which reconstructs images.
    These reconstructed frames along with the standard camera frames and IMU measurements allow the calibration toolbox to estimate intrinsics (K), and extrinsics (T) of the cameras as well as to the IMU.}
    \label{fig:approach}
    \vspace{-1.5ex}
\end{figure}
Camera calibration is an essential component of computer vision systems and has been thoroughly researched for decades \cite{Tsai87JRA}.
Typically, camera calibration methods extract corners from a known calibration pattern, detect the pattern and solve an optimization problem that optimizes for intrinsic and extrinsic parameters of the cameras.
This approach is nowadays widely used for standard frame-based cameras.
Unfortunately, this method cannot directly be used for event cameras.

Event cameras are asynchronous sensors that pose a paradigm shift in the way visual information is acquired.
In contrast to standard cameras that capture frames at regular intervals, event cameras report per-pixel brightness changes as a stream of asynchronous events.
Due to this property, image-based corner detection does not apply to event data such that standard calibration frameworks cannot be used.

Instead, recent calibration methods for event cameras rely on the usage of actively illuminated calibration patterns such as blinking LED patterns \cite{RPGCalib, Morales19} or electronic display devices \cite{Mueggler14iros,OrchardCalib, VLOCalib, PropheseeCalib}.
While blinking LED patterns can be used to calibrate event cameras, they cannot be reliably detected by standard cameras.
Hence, such an approach is not suitable for multiple camera calibration with both event and standard cameras. 
Electronic display devices on the other hand are not practical because it would require large screens for good coverage, especially for multiple camera calibration.
Furthermore, large screens are expensive, heavy, need additional power sources, and are cumbersome to move in front of a static camera setup.
Earlier academic versions of event cameras also captured monochrome images \cite{Brandli14iscas} which can be used for calibration \cite{Dubeau20ISMAR}. However, the latest industrial-grade event cameras do not provide image information anymore \cite{PropheseeGen4}. As a consequence, event-based calibration techniques are required.

Our proposed solution uses neural network-based image reconstruction and unlocks the full potential of traditional camera calibration framework.
The advantage of image reconstruction for event camera calibration is threefold.
First, we do not require an actively illuminated device to detect the calibration pattern.
Second, traditional calibration patterns can be detected by event and standard cameras alike.
This enables hybrid multiple camera calibration.
Third, the reconstructed images can be directly used by standard calibration frameworks, unlocking their potential for optimized calibration routines.

Our main contribution is the experimental validation of this approach for intrinsic, and extrinsic calibration of event cameras and hybrid setups involving standard cameras. We show that image reconstruction is a suitable tool to achieve accurate calibration performance in both synthetic and real-world experimental settings. 
\section{Related work}
In this section, we focus on related work and tools for achieving intrinsic and extrinsic camera calibration involving event cameras without active pixel sensor circuits \cite{Brandli14iscas}.

Open source toolboxes for intrinsic calibration of event cameras primarily use blinking LED patterns \cite{RPGCalib, Morales19} or blinking screens  \cite{Mueggler14iros,OrchardCalib, VLOCalib, PropheseeCalib} to extract the calibration pattern. Due to rapid change of illumination, the patterns can be detected even if it is static with respect to the camera frame. Once the pattern is extracted, standard optimization-based calibration back ends can be used. The main downsides of this approach is that it requires a custom built calibration board and that extrinsic event camera to standard camera calibration is not feasible.
%Another downside is that significant efforts are needed to denoise the event-accumulated frames as even small noise in the events can easily influence the blob detection leading to wrong calibration.

The closest work to ours \cite{MGehrig21RAL} applies image reconstruction for camera calibration. However, a detailed evaluation of their method and a comparison against other methods was never conducted, thus leaving the question open whether this method is accurate. 
In this work, we answer this question by building on these initial results and carefully evaluating them on both synthetic and real-world data from a variety of event cameras and lenses. 
We show that this calibration approach outperforms existing methods by a significant margin and yields consistently higher detection ratio than other approaches. We also show that the calibration method is more robust and can be used with a variety of lenses and cameras.
\section{Methodology}
In this section we describe the functionality of our event camera calibration method. 
First, in Sec. \ref{sec:method:event_data}, we introduce the working principle of the event camera and review the data structure measured by it. 
In Sec. \ref{sec:method:intrinsics} we present the general method of calibration of event camera through intensity reconstruction using E2VID\cite{Rebecq19pami}, an open-source neural network to reconstruct \rtwo{frames from events}, for camera intrinsics calibration.
\subsection{Event Data}\label{sec:method:event_data}
Event cameras have pixels that are independent and respond to changes in the
continuous log brightness signal $L(\mathbf{\textbf{u}_{k}}, t)$. An event $e_k=(x_k, y_k, t_k, p_k)$ is triggered when the magnitude of the log brightness at pixel
$u=(x_k, y_k)^T$ and time $t_k$ has changed by more than a threshold $C$ since
the last event at the same pixel. 
\begin{equation}
	\label{eq:generative_model}
	\Delta L(\mathbf{\textbf{u}_{k}}, t_k) = L(\mathbf{\textbf{u}_{k}}, t_k)-L(\mathbf{\textbf{u}_{k}}, t_k-\Delta t_k) \geq p_k C.
\end{equation}
Here, $\Delta t_k$ is the time since the last triggered event, $p_k\in\{-1,+1\}$
is the sign of the change, also called polarity of the event. Equation
\eqref{eq:generative_model} describes the generative event model for an ideal
sensor \cite{Gallego17pami, Gallego20pami}. 

\subsection{Camera Calibration}\label{sec:method:intrinsics}
%The task of camera intrinsics calibration has been traditionally performed with Zhang's Method \cite{Zhang00pami}. It requires several detections of a checkerboard with known row and column spacings from which it infers the camera calibration matrix $K$ and lens distortion parameters $D$ with
%\begin{equation}
%    K=
%    \begin{pmatrix}
%        f_x&0&u_0\\
%        0&f_y&v_0\\
%        0&0&1
%    \end{pmatrix}.
%\end{equation}
%Here $f_x$ and $f_y$ denote the focal length of the camera in pixels in the x and y direction, while $(u_0,v_0)$ denotes the optical center of the image plane. 
One of the fundamental building blocks of camera calibration is the detection of checkerboard corners, which for standard images is usually performed with corner detectors such as Harris\cite{Harris88} or Shi-Tomasi\cite{Shi94cvpr}. 
However, these detectors, originally designed for images, are not directly applicable to events due to their intrinsically asynchronous and sparse nature. For this reason we seek to convert the asynchronous and sparse event stream to dense images using the image reconstruction method in \cite{Rebecq19pami}. 
We then reuse the standard detectors presented above on these reconstructed images.

A high-level overview of our method is shown in figure \ref{fig:approach} and can be summarized in the following procedure:

\begin{enumerate}
    \item Divide event data in chunks of constant time duration. In our experiment we chose the duration to be $50$ milliseconds. This is a hyperparameter which may require modification to reach optimal performance. In case of intrinsic calibration of a single event camera, the time duration of these chunks does not have to be constant. For example, one could choose to define the chunks by the number of events within them.
    \item Reconstruct image from event data using E2VID \cite{Rebecq19pami}. If extrinsic calibration to global shutter cameras is performed, we reconstruct the image at the middle of the exposure time of the global shutter cameras.
    \item Prepare the image data for calibration according to the calibration toolbox of choice and proceed with the calibration. In our experiments we use the Kalibr calibration toolbox \cite{Furgale13iros}.
\end{enumerate}

In this procedure, we assume that all sensor data is synchronized in time.

%We describe an application scenario of hybrid multi-sensor configuration consisting of an event camera, a standard camera and a IMU.
%Our method uses frame-based calibration toolbox for calibrating the intrinsics of the event camera and camera ($K_e$ $K_c$) and the rigid body transformations from the event camera frame to IMU and transformation from the camera frame to IMU ($T\_event\_imu$ and $T\_camera\_imu$)

% \begin{figure}
%     \centering
%     \includegraphics{}
%     \caption{
%     Overview of our method. 
%     The asynchronous and sparse event stream is passed to the recurrent network E2VID\cite{Rebecq19pami} which reconstructs images at a fixed rate of 30 Hz.. 
%     We then use these images with the Kalibr toolbox\cite{Furgale13}.}
%     \label{fig:method}
% \end{figure}

\section{Experiments}\label{sec:experiment}
We \rtwo{assess} the performance of this method with simulated data and demonstrate the capability of this system for calibrating event cameras in real-world scenarios.
\rone{For all the experiments, we use the E2VID model provided by \cite{Rebecq19pami}.
We \textit{do not} perform any fine-tuning for the experiments.
Image reconstruction is performed with a fixed time window of $50$ milliseconds of events in all the experiments. }
\subsection{Intrinsic Calibration}\label{sec:experiment:intr}
\subsubsection{Simulation}\label{sec:experiment:simulation}
We simulate a camera moving in front of a calibration target to generate images and events.
%Simulation experiment
We use the state-of-the-art event simulation, ESIM\cite{Rebecq18corl}.
Given a camera trajectory, ESIM can simulate events, standard frames and IMU.
ESIM also allows incorporating distortion when simulating the events and frames.
We generate sequences with three distortion models: no-distortion, radial-tangential distortion model and equidistant distortion model.
The groundtruth distortion parameters used in this experiment are shown in \Tab \ref{tab:comparison:sim:params} as {$d1, d2, d3$ and $d4$}.
% Describe camera trajectory
We generate a camera trajectory in front of a calibration target, observing it from different viewpoints.
% Evaluation metric: Pattern detection accuracy
To evaluate the performance of E2VID in reconstructing frames with the calibration target, we use a detection accuracy metric.
We detect the pattern on all the E2VID frames and compute the number of images it could detect the pattern $p_{E2VID}$.
We also detect the pattern on all the ground truth frames and compute the number of detections $p_{GT}$.
We then calculate the success ratio as $\frac{p_{E2VID}}{p_{GT}}$
The calibration results for different calibration patterns and distortion types are summarized in the \Tab \ref{tab:comparison:sim}.

E2VID has a high success ratio for no distortion models, however the performance slightly decreases when significant amount of distortion is introduced for the checkerboard sequence.
Moreover, the detection accuracy decreases significantly for the AprilTag pattern as E2VID struggles to reconstruct the fine details of AprilTags, appearing smooth and leading to poor detection rate.
\rone{\Fig \ref{fig:experim:simulation:e2vid} shows exemplar image reconstructions for this experiment.}
We conclude that checkerboard patterns are preferable to AprilTag patterns for calibration with E2VID.

We also show the parameters that are estimated by E2VID reconstructions for the checkerboard and AprilTag pattern for all the distortion model in \Tab \ref{tab:comparison:sim:params}
The intrinsic parameters estimated by E2VID with the checkerboard calibration pattern are more accurate compared to AprilTag calibration, which is a direct consequence of the pattern detection accuracy.

\global\long\def\figWidth{0.24\linewidth}
\begin{figure}[t]
	\centering
\resizebox{1.1\columnwidth}{!}{
    \setlength{\tabcolsep}{2pt}
	\begin{tabular}{
	M{0.35cm}
	M{\figWidth}
	M{\figWidth}
	M{\figWidth}}
%		\midrule
		& No Distortion & Equidistant distortion & Radial-tangential distortion
		\\%\addlinespace[1ex]
		\rotatebox{90}{\makecell{Checkerboard}}
		&\frame{\includegraphics[width=\linewidth]{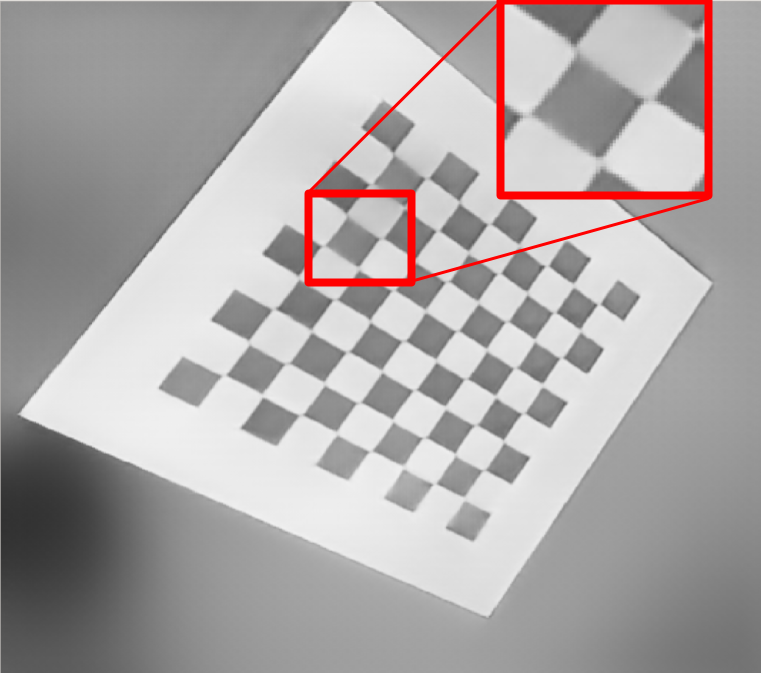}}
		&\frame{\includegraphics[width=\linewidth]{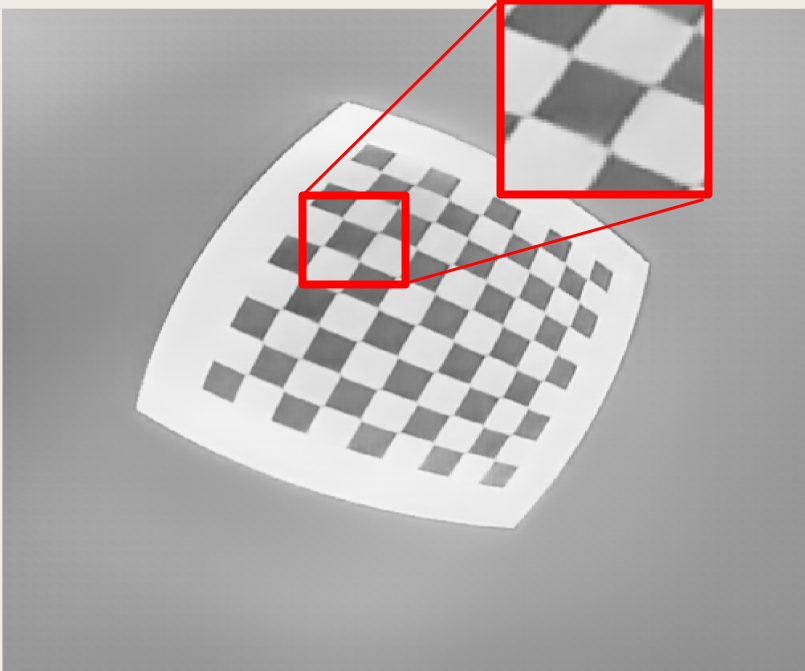}}
		&\frame{\includegraphics[width=\linewidth]{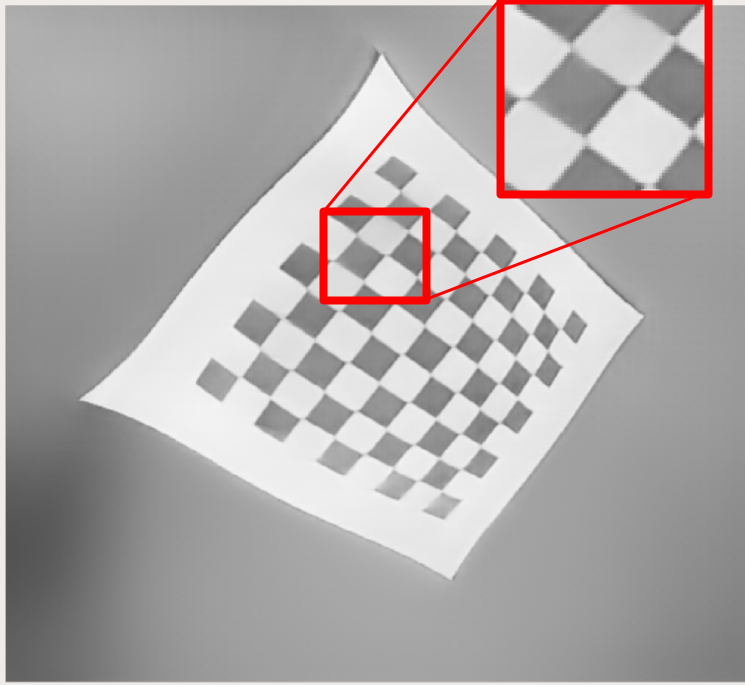}}
		\\%\addlinespace[1ex]
		\rotatebox{90}{\makecell{AprilTag}}
		&\frame{\includegraphics[width=\linewidth]{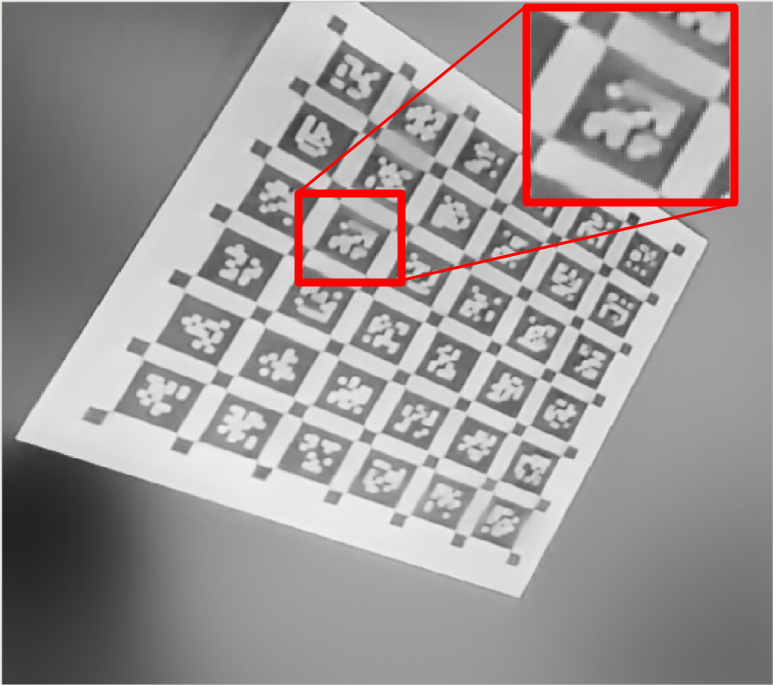}}
		&\frame{\includegraphics[width=\linewidth]{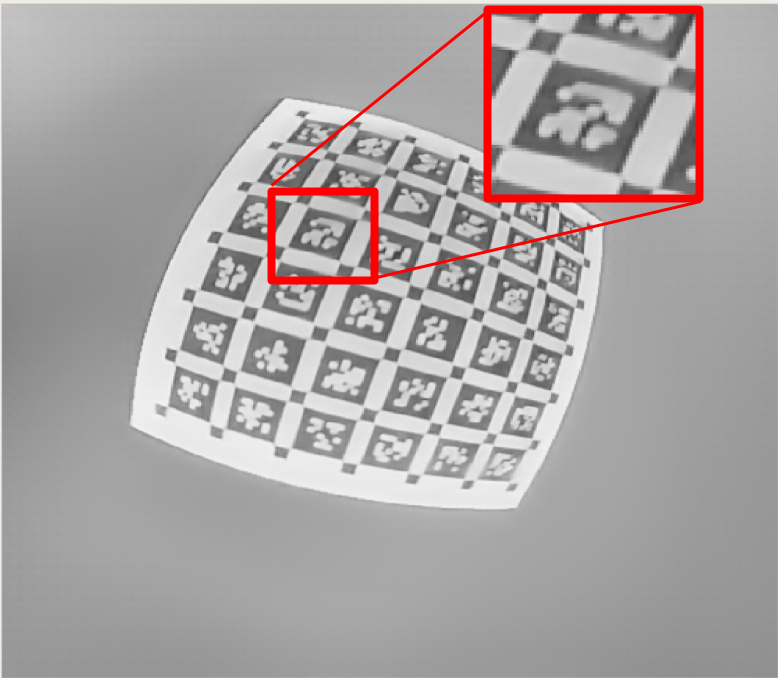}}
		&\frame{\includegraphics[width=\linewidth]{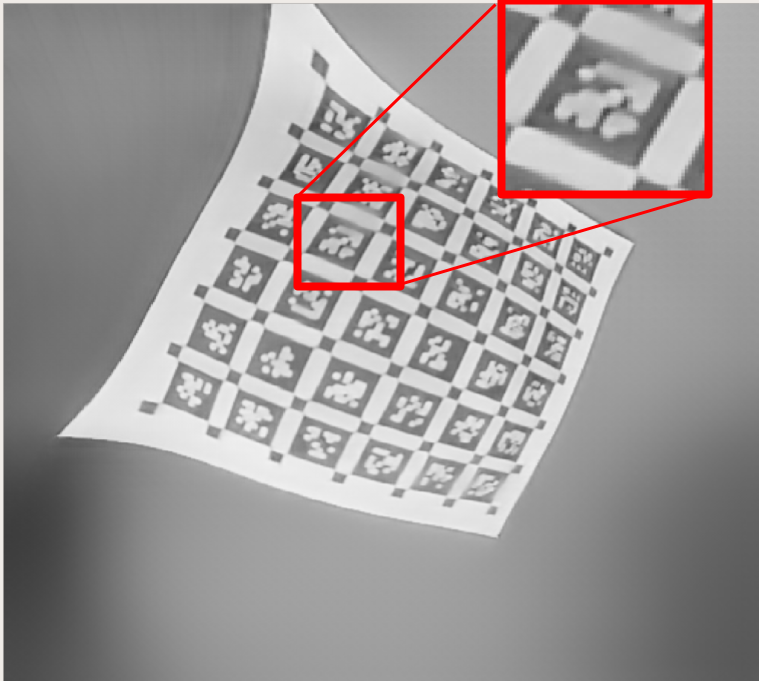}}
		\\%\addlinespace[1ex]
	\end{tabular}
	\vspace{-1ex}
} % resizebox
	\caption{E2VID reconstructed frames for simulation dataset for different distortion models.
	The reconstructions do not capture the fine details of april tag pattern.
	While the checkerboard reconstructions also suffer from some artifacts, these artifacts do not affect the corners.
	}
	\label{fig:experim:simulation:e2vid}
	%\vspace{-1ex}
\end{figure}
\begin{table}[t]
    \centering
    \begin{adjustbox}{max width=\linewidth}
    \setlength{\tabcolsep}{3pt}
    {\small
    \begin{tabular}{lrr}
        \toprule
        Pattern    & \multicolumn{1}{c}{Checkerboard}  & \multicolumn{1}{c}{AprilTag}   \\
        \midrule
                No distortion                & \textbf{1.0}     & \textbf{0.981} \\
                Equidistant distortion       & \textbf{0.981}   & 0.061 \\
                Radial-tangential distortion & \textbf{0.755}   &  0.329 \\
        \midrule
    \end{tabular}
    }
    \end{adjustbox}
    \vspace{-1ex}
    \caption{Pattern detection success ratio with different calibration targets, namely AprilTag and Checkerboard for different distortion models: no distortion, equidistant distortion, and radial-tangential.}
    \label{tab:comparison:sim}
    \vspace{-2ex}
\end{table}
\begin{table*}[t]
    \centering
    \begin{adjustbox}{max width=\linewidth}
    \setlength{\tabcolsep}{3pt}
    {\small
    \begin{tabular}{l|rrrr|rrrr}
        \toprule
        Parameters      & \multicolumn{1}{c}{fx} & \multicolumn{1}{c}{fy}   & \multicolumn{1}{c}{cx}   & \multicolumn{1}{c}{cy}   & \multicolumn{1}{c}{d1}   & \multicolumn{1}{c}{d2}   & \multicolumn{1}{c}{d3}   & \multicolumn{1}{c}{d4} \\
        \midrule
                No distortion (AprilTag) & 177.35 &	176.75 &	242.97 &	249.84 &	-0.004 &	0 &	0 &	0\\
                No distortion (Checkerboard) & 201.82 &	201.99 &	252.14 &	250.86 &	-0.003	& 0	& 0 &	0\\
                GT    & 200 & 200 & 250 & 250 &	0 & 0 &	0 &	0 \\\midrule
                Equidistant distortion (AprilTag)       & 167.05 &	166.98 &	250.69 &	249.95 &	0.029 &	0.114 &	-0.161 &	0.0716\\
                Equidistant distortion (Checkerboard)       & 205.11 &	204.91	& 250.05 &	249.91 &	-0.073 &	0.055 &	-0.085 &	0.056 \\
                GT & 200 &	200 &	250 &	250 &	-0.051 &	0.046 &	-0.083 &	0.056\\\midrule
                Radial-tangential distortion (AprilTag) & 187.26 &	186.99 &	248.59 &	250.9 &	-0.323 &	0.132 &	-0.0024 &	0.002\\
                Radial-tangential distortion (Checkerboard) & 187.8 &	188.16 &	250.24 &	249.86 &	-0.336 &	0.145 &	-0.002 &	-0.001\\
                GT & 200&	200 &	250 &	250 &	-0.383 &	0.189 &	-0.001 &	-0.001\\
        \midrule
    \end{tabular}
    }
    \end{adjustbox}
    \vspace{-1ex}
    \caption{Kalibr camera intrisics parameters estimation with synthetic data of moving calibration target of checkerboard and AprilTag pattern. Camera calibration is performed on the reconstructed frames from E2VID. fx and fy are the focal lengths, cx and cy the principal point coordinates, and d1 to d4 the distortion parameters.}
    \label{tab:comparison:sim:params}
    \vspace{-2ex}
\end{table*}

\subsubsection{Real-world Intrinsic Calibration} \label{sec:experiment:intrinsic}

\begin{figure*}[t]
    \centering
    \includegraphics[width=0.32\linewidth]{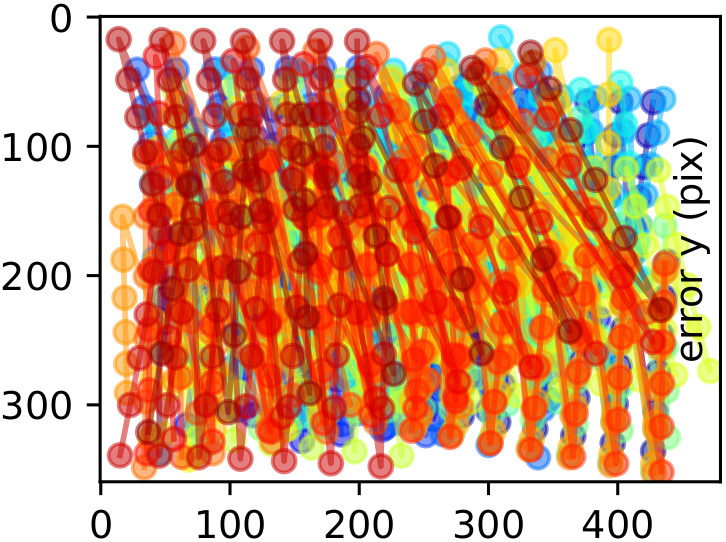}
    \includegraphics[width=0.32\linewidth]{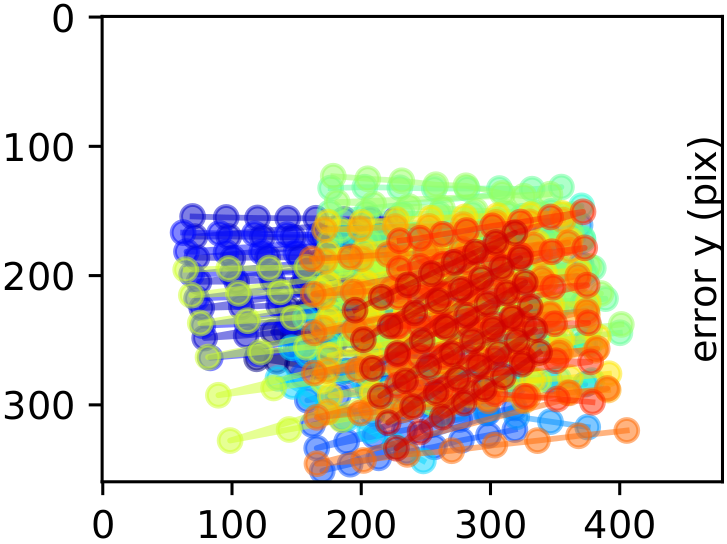}
    \includegraphics[width=0.32\linewidth]{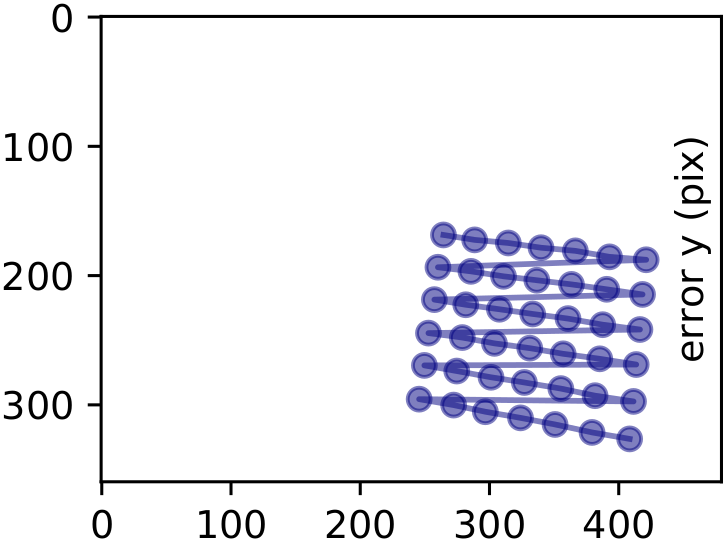}
    \caption{Coverage of detected patterns by Kalibr in the image plane of the Gen3 ATIS event camera. From left to right: E2VID, LCD, and EM frames. In case of E2VID reconstructions, the whole image plane is covered with detected patterns while this is not the case for LCD input or EM frames. Good coverage is key for achieving accurate calibration.}
    \label{fig:kalibr_coverage_gen3}
\end{figure*}

\begin{figure}[t]
    \centering
    \includegraphics[height=3cm]{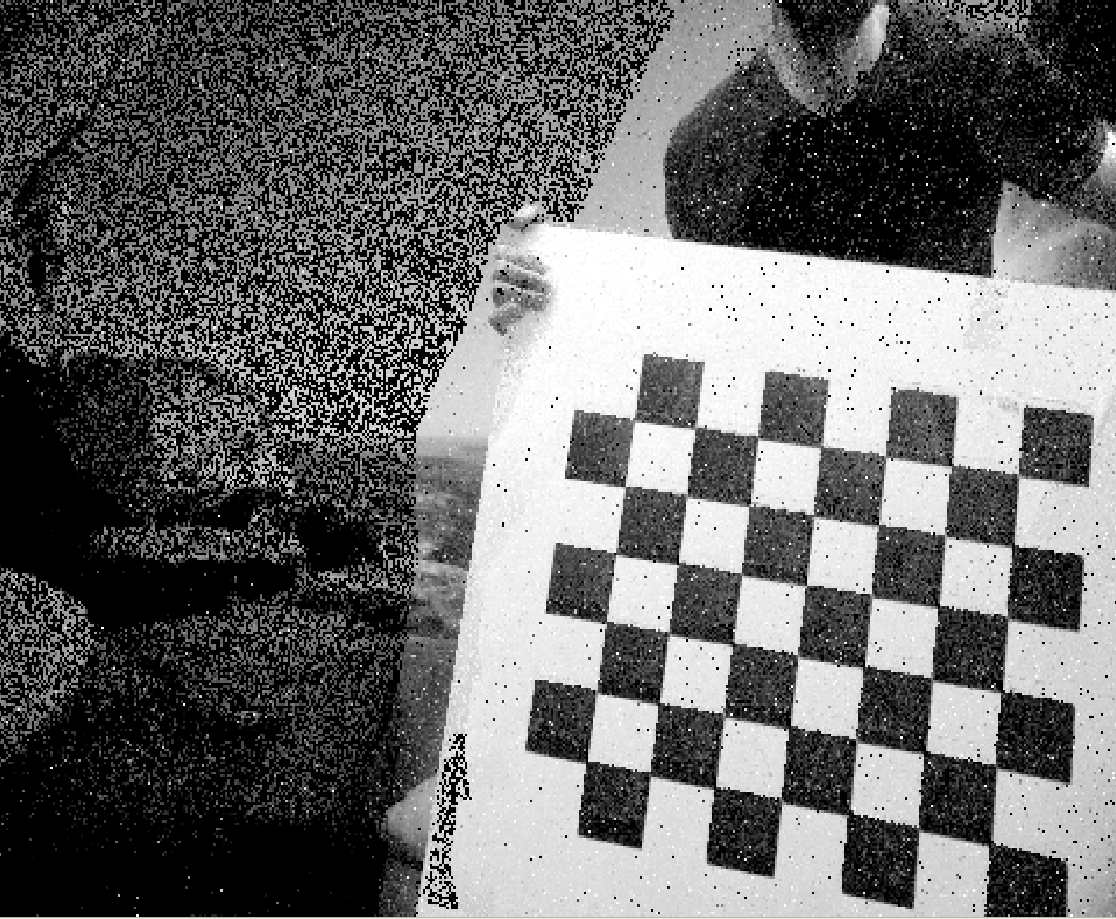}
    \includegraphics[height=3cm]{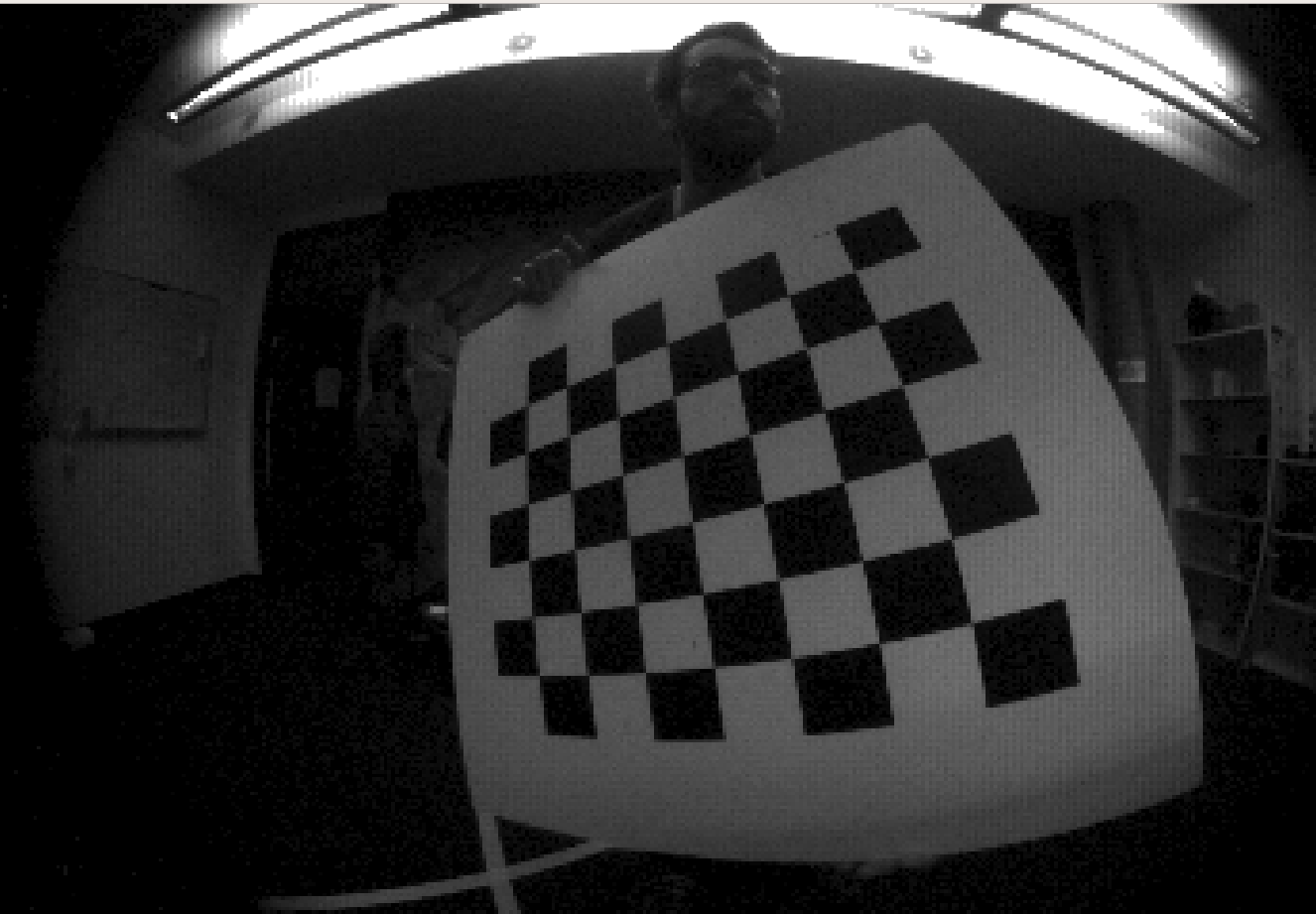}
    \caption{Comparison of an EM frame from the Gen3 ATIS camera and a frame from the DAVIS346. The EM frames are noisier compared to DAVIS frames and pose a challenge for detecting calibration patterns.}
    \label{fig:em_frames_vs_davis_frames}
\end{figure}

We now compare our approach with previously used baselines for calibration such as the blinking LED board and the LCD screen.
Here, we use 4 different event camera sensors namely: DAVIS346 \cite{Brandli14iscas} (resolution $346 \times 260$), Samsung Gen3 \cite{Son17isscc} (resolution $640 \times 480$), Prophesee Gen3 ATIS \cite{Posch10isscc, propheseeevk} (resolution $480 \times 360$) and Prophesee Gen4 \cite{Finateu20isscc} (resolution $1280 \times 720$).
The sensors cover a wide range of camera resolutions and distortions.
We collect calibration sequence from all the 4 sensors for three different calibration methods: (\emph{i}) blinking LED pattern, we create a $5 \times 5$ grid of LED lights flickering at $500$ Hz,
(\emph{ii}) LCD screen, we use a LCD monitor to display a checkerboard pattern flickering at $60$ Hz
(\emph{iii}) Checkerboard, we use a traditional checkerboard plane which moves in front of the camera.
Apart from the events, DAVIS346 and Prophesee Gen3 sensor also provide the grayscale frame.
\rone{An important point to note here is that the grayscale frames from DAVIS346 are global shutter frames whereas the prophesee grayscale frames are generated from exposure measurement (EM) events which encode the intensity of the event.
For clarification, we refer to these frames as EM frames.
Since EM frames are formed by events these are typically noisy compared to DAVIS346 frames. Figure \ref{fig:em_frames_vs_davis_frames} shows this qualitatively.}

We use the calibration on grayscale frame as a baseline to compare the accuracy of the event-based calibration methods only for the DAVIS346 and Prophesee Gen3 sensor.
While the LCD screen and the checkerboard sequences can be used with Kalibr toolbox, the blinking LED pattern uses the OpenCV functions for blob detection followed by the \emph{cameraCalibrate} function.

\begin{figure}[t]
    \centering
    \includegraphics[height=3cm]{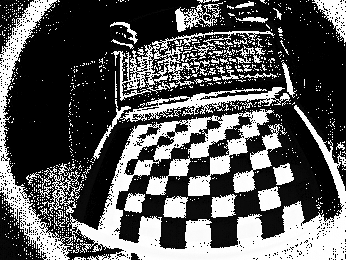}
    \includegraphics[height=3cm]{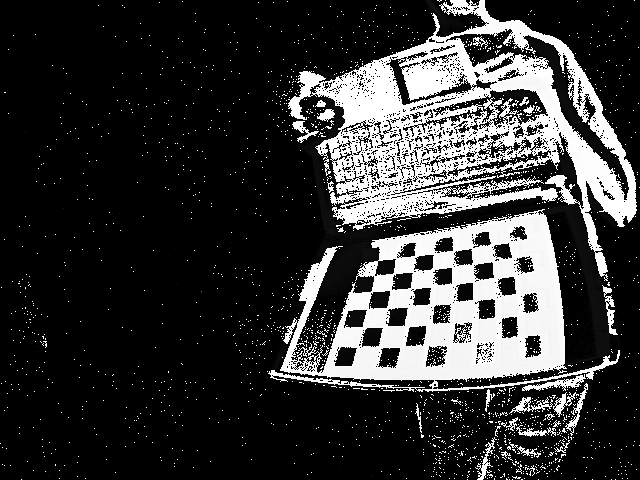}
    \caption{Event frames from DAVIS346 and Samsung Gen3 for calibration with LCD screen.}
    \label{fig:lcd}
\end{figure}

%Detection accuracy
\rtwo{Table \ref{tab:comparison:detection} reports the ratio of number of frames where the pattern was detected by the total number of frames in the sequence. This is done for each calibration sequence with the different calibration patterns.}
This metric provides an estimate of how much of the image plane is covered by the detected frames; an example of this is shown in figure \ref{fig:kalibr_coverage_gen3} for the Gen3 ATIS event camera.
A low pattern detection score would imply a lower coverage but it does not necessarily imply high reprojection error. Instead, with low coverage, the intrinsic parameters may not be valid for the whole image area and lead to suboptimal performance in real-world applications.
We observe that the LCD screen performs the worst while the blinking LED pattern performs slightly better.
\rone{The LCD screen performs worse because of the low detection ratio. 
This low detection ratio is due to (\emph{i}) the movement of the LCD screen which causes motion blur in the frames and  (\emph{ii}) noisy events triggered by the blinking of the LCD screen.
The \Fig \ref{fig:lcd} shows examples of event frames from the DAVIS346 and Samsung Gen3 where the pattern is most visible.}
In contrast, our method, outperforms the baselines achieving a performance closest to the frame-based method (for DAVIS346 and Prophesee Gen3).

% Reprojection error
The reprojection error can be seen as a metric that informs about the accuracy of the detection front end. If the extracted keypoints are always very close to the ideal keypoints, this metric is usually low. However, as mentioned in the previous paragraph, this metric can also be low when the pattern is only detected in a small portion of the image area. Hence, low reprojection error is a necessary condition for successful calibration, not a sufficient one. 
Table \ref{tab:comparison:reproj} compares the root mean square (RMS) reprojection error after the calibration procedure for all the calibration pattern sequences.
We observe that the RMS is consistently low for our proposed method. Only DAVIS frames report a lower reprojection error overall. Compared to the E2VID and checkerboard combination, the LCD screen RMS reprojection error is at least two times higher or even fails for the DAVIS and Samsung cameras. \rtwo{The blinking LED calibration pattern is the least accurate in our experiments.} To summarize, the E2VID calibration is successful and accurate for all tested cameras.

\subsection{Multi-Sensor Extrincis Calibration}\label{sec:experiment:multi_cam}
\begin{table*}[t]
    \centering
    \begin{adjustbox}{max width=\linewidth}
    \setlength{\tabcolsep}{3pt}
    {\small
    \begin{tabular}{llrrrr}
        \toprule
        Sensor  &  Modality & \multicolumn{1}{c}{Davis346}  & \multicolumn{1}{c}{Samsung Gen3}& \multicolumn{1}{c}{Prophesee Gen3} & \multicolumn{1}{c}{Prophesee Gen4}   \\
        \midrule
                   Blinking LED         & \emph{E}&  0.04 & 0.06 & 0.035 & 0.0006 \\
                   LCD screen          & \emph{E} &  0.003   & 0.016 & 0.0049  & 0.0026\\
                %   Checkerboard(Event frames) & \emph{E} &  &  & &  \\
                   Checkerboard (E2VID)  & \emph{E} &  0.47 & \textbf{0.65} & \textbf{0.155} & \textbf{0.217} \\\midrule
                   Checkerboard (Frames) & \emph{I} &  \textbf{0.61} & -   & 0.0434 & - \\
        \bottomrule
    \end{tabular}
    }
    \end{adjustbox}
    \vspace{-1ex}
    \caption{Pattern detection ratio for calibration sequences across different sensor resolutions and distortions. The input modalities \emph{E} and \emph{I} represent events and images respectively that are used for the calibration method.
    Where \emph{I} is not available it is marked as -.}
    % Format: yes if successful reprojection error else "-"(number of images where corners were detected.)}
    \label{tab:comparison:detection}
    \vspace{-2ex}
\end{table*}

\begin{table*}[t]
    \centering
    \begin{adjustbox}{max width=\linewidth}
    \setlength{\tabcolsep}{3pt}
    {\small
    \begin{tabular}{llrrrr}
        \toprule
        Sensor  &  Modality & \multicolumn{1}{c}{Davis346}  & \multicolumn{1}{c}{Samsung Gen3}& \multicolumn{1}{c}{Prophesee Gen3} & \multicolumn{1}{c}{Prophesee Gen4}   \\
        \midrule
                   Blinking LED & \emph{E} & 4.38 & 6.20 & 10.93 & 13..91 \\
                   LCD screen          & \emph{E} &  x   & x & 0.79 & 0.81\\
                %   Checkerboard(Event frames) & \emph{E} &  &  & &  \\
                   Checkerboard (E2VID)  & \emph{E} & 0.29 & \textbf{0.21} & \textbf{0.24} & \textbf{0.45} \\\midrule
                   Checkerboard (Frames) & \emph{I} &  \textbf{0.17} & -   &  0.50 & - \\
        \bottomrule
    \end{tabular}
    }
    \end{adjustbox}
    \vspace{-1ex}
    \caption{Root mean square (RMS) reprojection error after the calibration procedure of Kalibr. 
    The input modalities \emph{E} and \emph{I} represent events and images respectively that are used for the calibration method. 
    Where \emph{I} is not available it is marked as -.
    We report the RMS reprojection error where Kalibr is successful, otherwise it is marked as x. 
    Lowest (best) RMSE in bold.
    }

    \label{tab:comparison:reproj}
    \vspace{-2ex}
\end{table*}

% second table for reprojection error (std devaiation) for LCD and checkerboard. ->table_reprojection_error

% \input{floats/table_focal_lengths}
\begin{figure}[t]
    \centering
    \includegraphics[trim={1cm 3.1cm 3cm 0.8cm},clip,width=0.7\linewidth]{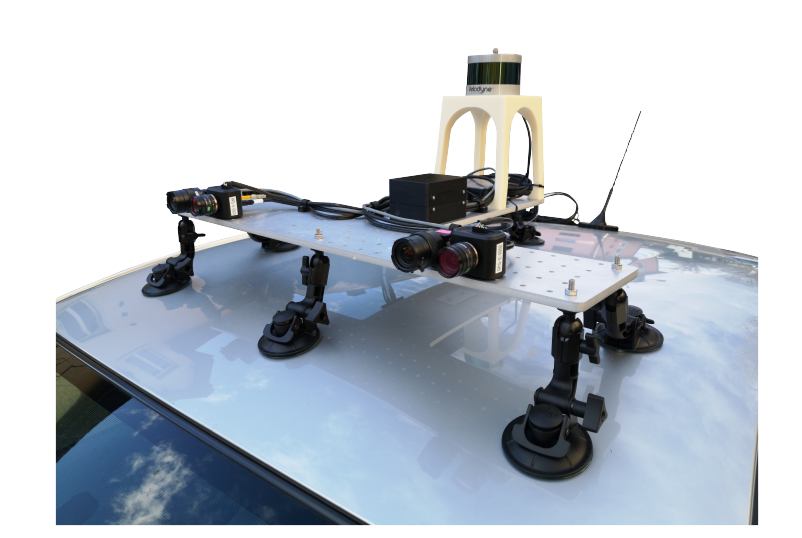}
    \vspace{-1ex}
    \caption{Hybrid multiple camera setup consisting of 2 Prophesee event cameras, 2 FLIR global shutter cameras, which is used for testing calibration consistency.}
    \label{fig:dsec_Setup}
    \vspace{-1.5ex}
\end{figure}
The advantage of converting events to frames and using them for calibration is that this allows us to use standard calibration toolboxes designed for calibrating a vast array of sensors with respect to a standard frame-based camera.
We demonstrate this by calibrating the extrinsic and intrinsic parameters of an event camera in combination with (\emph{i}) a standard-frame-based camera and (\emph{ii}) an inertial measurement unit (IMU).

\subsubsection{Event Cameras and Standard Camera Calibration}
\begin{figure}[t]
    \centering
    \includegraphics[width=0.95\linewidth]{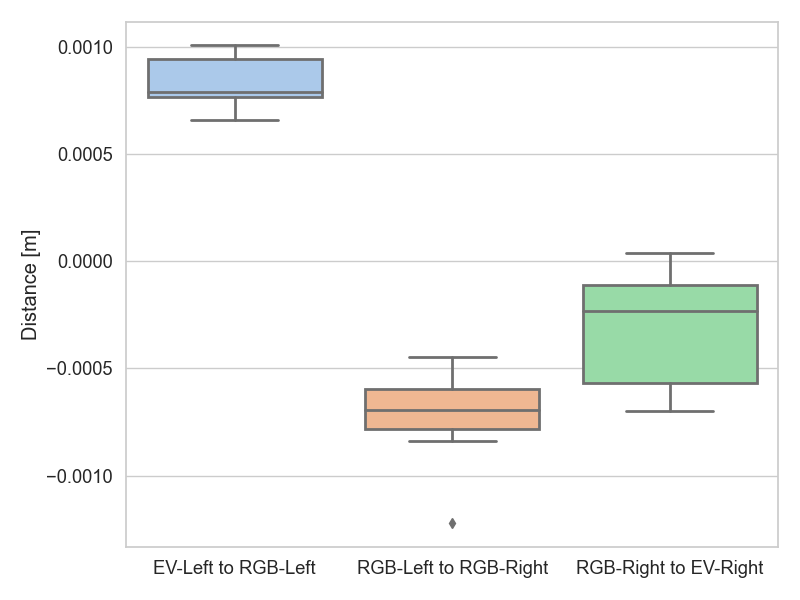}
    \caption{Boxplots visualizing the deviation from the estimated translation between cameras to the translation retrieved from the CAD model.}
    \label{fig:dsec_boxplot}
\end{figure}
For this experiment, we use 4 hardware synchronized cameras consisting of 2 global shutter RGB cameras and 2 event cameras in a dual-stereo setup, depicted in \Fig \ref{fig:dsec_Setup}.
% \rone{All the cameras are synchronized }
The two event cameras are at the rightmost and leftmost location of the setup and consist of VGA resolution event cameras.
According to our CAD model the two standard cameras have a baseline of 0.51 meters.
In this setting, we examine the repeatability and accuracy of multiple hybrid camera calibration with our proposed approach. To assess the repeatability of the approach, we calibrate the setup 11 times over the course of 1 month while it is used for data collection. During this time, the translation between the cameras remains approximately the same while the rotation and intrinsic slightly change due to manipulation of the setup. \Fig \ref{fig:dsec_boxplot} shows the boxplot of the translational deviation from the CAD model and estimated translation by our proposed framework. Our CAD model is accurate up to around 5 millimeter of translation. The errors are all well within this limit, which shows that the extrinsic calibration is accurate within the tolerance of our CAD model. Finally, the narrow error bars suggest that the calibration could potentially reach sub-millimeter accuracy. However, a conclusion in that regard requires more controlled experiments with highly accurate CAD models.

\subsubsection{Event Camera and IMU Calibration}\label{sec:experiment:cam_imu}
Integrating IMU inertial information along with visual information advanced the progress of visual odometry.
\cite{Rosinol18ral} showed the advantage of combining events with IMU and frames for visual-odometry in high dynamic range and high speed motion.
Combining these sensors requires a knowledge of the rigid body transformations between them.
In this section, we show that by using E2VID for events, we are able to calibrate an event camera with an IMU using standard calibration toolbox Kalibr \cite{Furgale13iros}.
We compare the result of Kalibr toolbox for camera-imu calibration on two methods (\emph{i}) using ATIS frames and IMU from the Prophesee Gen3 \cite{Posch11ssc} and  (\emph{ii}) using E2VID reconstructed frames and IMU from the Prophesee Gen3 \cite{Posch11ssc}
The root-mean-squared error between the translation vectors estimated by these two methods is $0.74$ \SI{}{\centi\meter}, indicating that by using E2VID in our calibration framework, we are able to accurately calibrate our systems using \textit{only events}.
\section{Conclusion}
In this paper, we propose a framework for calibrating an event camera using image reconstruction.
Our method does not require a special calibration target like blinking pattern or external monitor screens, previously used for calibrating event cameras.
By converting events to frames using E2VID\cite{Rebecq19pami} for calibration, we unlock the potential of event cameras to be used in multi-sensor configurations for applications like autonomous driving, augmented/mixed reality and robotics.
Our experiments both simulation and real-world indicate that calibration through image-reconstruction is accurate under prevalent distortion models and wide variety of possible parameter sets.

% Maybe we can use the following for an application of another paper. So leaving it away.
%\section{Future Work}
%We showed that image reconstructions by E2VID lead to accurate calibration results. Still, E2VID was not primarily designed for event camera calibration. Possible improvements to the current methodology include specialized image reconstruction for the purpose of calibration. This can be achieved by altering the loss to achieve more accurate image gradients and removing temporal consistency loss that was introduced to alleviate video flickering. Furthermore, E2VID training data does not contain calibration patterns which are the sole interest of camera calibration. Finally, E2VID requires an input representation from a certain duration of event data. This introduces a hyperparameter that may lead to suboptimal results. An image reconstruction method that is largely independent of the choice of temporal discretization could further improve the calibration accuracy and simplify the workflow for event camera calibration.

\section{Acknowledgements}
 This work was supported by SONY R\&D Center Europe, the National Centre of Competence in Research (NCCR) Robotics through the Swiss National Science Foundation (SNSF), and the European Research Council (ERC) under grant agreement No. 864042 (AGILEFLIGHT).
{\small
\balance
\bibliographystyle{ieeetr_fullname}
\bibliography{all}

\begin{thebibliography}{10}\itemsep=-1pt

\bibitem{Rebecq19pami}
Henri Rebecq, Ren{\'{e}} Ranftl, Vladlen Koltun, and Davide Scaramuzza, ``High
  speed and high dynamic range video with an event camera,'' {\em {IEEE} Trans.
  Pattern Anal. Mach. Intell.}, 2019.

\bibitem{Tsai87JRA}
R. {Tsai}, ``A versatile camera calibration technique for high-accuracy 3d
  machine vision metrology using off-the-shelf tv cameras and lenses,'' {\em
  IEEE Journal on Robotics and Automation}, vol.~3, no.~4, pp.~323--344, 1987.

\bibitem{RPGCalib}
``Calibration toolbox by rpg group.''
  \url{https://github.com/uzh-rpg/rpg_dvs_ros/tree/master/dvs_calibration},
  2014.

\bibitem{Morales19}
M.~J. {Domínguez-Morales}, Á. {Jiménez-Fernández}, G. {Jiménez-Moreno}, C.
  {Conde}, E. {Cabello}, and A. {Linares-Barranco}, ``Bio-inspired stereo
  vision calibration for dynamic vision sensors,'' {\em IEEE Access}, vol.~7,
  pp.~138415--138425, 2019.

\bibitem{Mueggler14iros}
Elias Mueggler, Basil Huber, and Davide Scaramuzza, ``Event-based, 6-{DOF} pose
  tracking for high-speed maneuvers,'' in {\em IEEE/RSJ Int. Conf. Intell.
  Robot. Syst. (IROS)}, pp.~2761--2768, 2014.

\bibitem{OrchardCalib}
``Calibration toolbox by garrick orchard.''
  \url{https://github.com/gorchard/DVScalibration}, 2015.

\bibitem{VLOCalib}
``Calibration toolbox by vlo group.''
  \url{https://github.com/VLOGroup/dvs-calibration}, 2017.

\bibitem{PropheseeCalib}
``Calibration toolbox by prophesee.''
  \url{https://docs.prophesee.ai/metavision_sdk/modules/calibration/guides/intrinsics.html},
  2020.

\bibitem{Brandli14iscas}
Christian Brandli, Lorenz Muller, and Tobi Delbruck, ``Real-time, high-speed
  video decompression using a frame- and event-based {DAVIS} sensor,'' in {\em
  {IEEE} Int. Symp. Circuits Syst. (ISCAS)}, pp.~686--689, 2014.

\bibitem{Dubeau20ISMAR}
E. {Dubeau}, M. {Garon}, B. {Debaque}, R. d. {Charette}, and J.~F. {Lalonde},
  ``Rgb-d-e: Event camera calibration for fast 6-dof object tracking,'' in {\em
  2020 IEEE International Symposium on Mixed and Augmented Reality (ISMAR)},
  pp.~127--135, 2020.

\bibitem{PropheseeGen4}
T. {Finateu}, A. {Niwa}, D. {Matolin}, K. {Tsuchimoto}, A. {Mascheroni}, E.
  {Reynaud}, P. {Mostafalu}, F. {Brady}, L. {Chotard}, F. {LeGoff}, H.
  {Takahashi}, H. {Wakabayashi}, Y. {Oike}, and C. {Posch}, ``5.10 a 1280×720
  back-illuminated stacked temporal contrast event-based vision sensor with
  4.86µm pixels, 1.066geps readout, programmable event-rate controller and
  compressive data-formatting pipeline,'' in {\em 2020 IEEE International
  Solid- State Circuits Conference - (ISSCC)}, pp.~112--114, 2020.

\bibitem{MGehrig21RAL}
M. {Gehrig}, W. {Aarents}, D. {Gehrig}, and D. {Scaramuzza}, ``Dsec: A stereo
  event camera dataset for driving scenarios,'' {\em IEEE Robotics and
  Automation Letters}, pp.~1--1, 2021.

\bibitem{Gallego17pami}
Guillermo Gallego, Jon E.~A. Lund, Elias Mueggler, Henri Rebecq, Tobi Delbruck,
  and Davide Scaramuzza, ``Event-based, 6-{DOF} camera tracking from
  photometric depth maps,'' {\em {IEEE} Trans. Pattern Anal. Mach. Intell.},
  vol.~40, pp.~2402--2412, Oct. 2018.

\bibitem{Gallego20pami}
Guillermo Gallego, Tobi Delbruck, Garrick Orchard, Chiara Bartolozzi, Brian
  Taba, Andrea Censi, Stefan Leutenegger, Andrew Davison, J{\"o}rg Conradt,
  Kostas Daniilidis, and Davide Scaramuzza, ``Event-based vision: A survey,''
  {\em {IEEE} Trans. Pattern Anal. Mach. Intell.}, 2020.

\bibitem{Harris88}
Chris Harris and Mike Stephens, ``A combined corner and edge detector,'' in
  {\em Proc. Fourth Alvey Vision Conf.}, vol.~15, pp.~147--151, 1988.

\bibitem{Shi94cvpr}
Jianbo Shi and Carlo Tomasi, ``Good features to track,'' in {\em {IEEE} Conf.
  Comput. Vis. Pattern Recog. (CVPR)}, pp.~593--600, 1994.

\bibitem{Furgale13iros}
P. Furgale, J. Rehder, and R. Siegwart, ``Unified temporal and spatial
  calibration for multi-sensor systems,'' in {\em IEEE/RSJ Int. Conf. Intell.
  Robot. Syst. (IROS)}, 2013.

\bibitem{Rebecq18corl}
Henri Rebecq, Daniel Gehrig, and Davide Scaramuzza, ``{ESIM}: an open event
  camera simulator,'' in {\em Conf. on Robotics Learning (CoRL)}, 2018.

\bibitem{Son17isscc}
Bongki Son, Yunjae Suh, Sungho Kim, Heejae Jung, Jun-Seok Kim, Changwoo Shin,
  Keunju Park, Kyoobin Lee, Jinman Park, Jooyeon Woo, Yohan Roh, Hyunku Lee,
  Yibing Wang, Ilia Ovsiannikov, and Hyunsurk Ryu, ``A 640x480 dynamic vision
  sensor with a 9$\mu$m pixel and {300Meps} address-event representation,'' in
  {\em {IEEE} Intl. Solid-State Circuits Conf. (ISSCC)}, 2017.

\bibitem{Posch10isscc}
Christoph Posch, Daniel Matolin, and Rainer Wohlgenannt, ``A {QVGA} 143{dB}
  dynamic range asynchronous address-event {PWM} dynamic image sensor with
  lossless pixel-level video compression,'' in {\em {IEEE} Intl. Solid-State
  Circuits Conf. (ISSCC)}, pp.~400--401, 2010.

\bibitem{propheseeevk}
{Prophesee Evaluation Kits}. \url{https://www.prophesee.ai/event-based-evk/},
  2020.

\bibitem{Finateu20isscc}
Thomas Finateu, Atsumi Niwa, Daniel Matolin, Koya Tsuchimoto, Andrea
  Mascheroni, Etienne Reynaud, Pooria Mostafalu, Frederick Brady, Ludovic
  Chotard, Florian LeGoff, Hirotsugu Takahashi, Hayato Wakabayashi, Yusuke
  Oike, and Christoph Posch, ``A 1280x720 back-illuminated stacked temporal
  contrast event-based vision sensor with 4.86$\mu$m pixels, 1.066geps readout,
  programmable event-rate controller and compressive data-formatting
  pipeline,'' in {\em {IEEE} Intl. Solid-State Circuits Conf. (ISSCC)}, 2020.

\bibitem{Rosinol18ral}
Antoni {Rosinol Vidal}, Henri Rebecq, Timo Horstschaefer, and Davide
  Scaramuzza, ``Ultimate {SLAM}? combining events, images, and {IMU} for robust
  visual {SLAM} in {HDR} and high speed scenarios,'' {\em {IEEE} Robot. Autom.
  Lett.}, vol.~3, pp.~994--1001, Apr. 2018.

\bibitem{Posch11ssc}
Christoph Posch, Daniel Matolin, and Rainer Wohlgenannt, ``A {QVGA} 143 {dB}
  dynamic range frame-free {PWM} image sensor with lossless pixel-level video
  compression and time-domain {CDS},'' {\em {IEEE} J. Solid-State Circuits},
  vol.~46, pp.~259--275, Jan. 2011.

\end{thebibliography}
}

\end{document}